# A Neural Network based Approach for Predicting Customer Churn in Cellular Network Services

Anuj Sharma
Information Systems Area
Indian Institute of Management,
Indore, India

Dr. Prabin Kumar Panigrahi
Information Systems Area
Indian Institute of Management,
Indore, India

## ABSTRACT
Marketing literature states that it is more costly to engage a new customer than to retain an existing loyal customer. Churn prediction models are developed by academics and practitioners to effectively manage and control customer churn in order to retain existing customers. As churn management is an important activity for companies to retain loyal customers, the ability to correctly predict customer churn is necessary. As the cellular network services market becoming more competitive, customer churn management has become a crucial task for mobile communication operators. This paper proposes a neural network (NN) based approach to predict customer churn in subscription of cellular wireless services. The results of experiments indicate that neural network based approach can predict customer churn with accuracy more than 92%. Further, it was observed that medium sized NNs perform best for the customer churn prediction when different neural network's topologies were experimented.

## 1. INTRODUCTION
Enterprises in the competitive market mainly rely on the incessant profits which come from existing loyal customers. Therefore, customer relationship management (CRM) always concentrates on loyal customers that are the most fertile and reliable source of data for managerial decision making. This data reflects customers' actual individual product or service consuming behavior. This kind of behavioral data can be used to evaluate customers' potential life time value [1], to assess the risk that they will stop paying their bills or will stop using any products or services, and to anticipate their future needs [2].

Today, customer relationship management (CRM) systems are replacing traditional mass marketing strategies by selective or personalized marketing practices [3]. These selective marketing practices involve identifying a sub-set of existing customers that are likely to stop using products or services of the company (churn). As existing customer's churning will likely to result in the loss of businesses and thus decline in profit, churn prediction has received increasing consideration in the consumer marketing and management research literature over the past few years. In addition, literature suggests that a small change in the retention rate can result in significant impact on business [4].

In order to effectively control customer churn, it is important to build a more effective and accurate customer churn prediction model. Statistical and data mining techniques have been utilized to construct the churn prediction models. The data mining techniques can be used to discover interesting patterns or relationships in the data, and predict or classify the behavior by fitting a model based on available data. In other words, it is an interdisciplinary area with a general objective of predicting outcomes and employing sophisticated data processing algorithms to discover mainly hidden patterns, associations, anomalies, and/or structure from extensive data stored in data warehouses or other information repositories [5].

There are several data mining techniques that are proposed to predict potential customers that are most likely to churn. Among the popular techniques to predict customer churn are: neural networks, support vector machines and logistic regression models [6]. Data mining research literature suggests that for non-parametric datasets, machine learning techniques, such as neural networks, often outperform traditional statistical and structurally restrictive techniques such as linear and quadratic discriminant analysis approaches [7].

As we consider from business intelligence perspective, churn management processes under the customer relationship management (CRM) framework consists of two major analytical modeling tasks. First task is predicting those customers who are about to churn and second task is assessing the most effective way that an service operator can react that includes providing special promotion programs to customer or simply do nothing for customer retention.

This research supports the former task, means it intends to illustrate how to apply data mining techniques to assist telecom churn management. This paper suggests artificial neural network (ANN) technique to find a best model from stored customer data to predict churn and to prevent the customer's turnover. Like this way, the cellular service operators can enhance the competitive edge.

This paper proposes a neural network (NN) based approach to predict customer churn in subscription of cellular wireless services. The rest of the paper is organized as follows. Section 2 reviews the current literature, related to customer churn and different data mining techniques used to predict churn in different studies. Section 3 describes the current research methodology, and Section 4 presents the experimental results on real time dataset. Finally, conclusion is provided in Section 5.

## 2. LITERATURE REVIEW
### 2.1 Customer Churn Management
As the market becoming more competitive, many companies have started realization of the importance of customer-oriented business strategy for sustaining their competitive edge and maintaining a stable profit level at top line and bottom line. That is, companies mainly rely on the stable income which comes from loyal customers. However, creating new customers and retaining loyal customers is difficult and costly. As new customer account is setup different expenses like credit searching, advertising and promotional expenses are occurred. These expenses are several times greater than cost of efforts that might enable the firms to retain a customer [8]. So, it has developed into an industry-wide belief that the best core marketing strategy for the future is to retain existing customers and avoid customer churn [9].

Marketing research literature has noted that 'customer churn' is a term used in the cellular mobile telecom service industry to indicate the customer movement from one provider to another, and 'churn management' is a term that describes an operator's process to retain profitable customers [10]. Churn is also called attrition and often used to indicate a customer leaving the service of one company in favor of another company. Similarly, the term churn management in the cellular network services industry is used to describe the practices of securing the most important customers for a company [11]. In essence, effective customer management presumes an ability to forecast the customer decision to shift from one service provider to another. Customer management also presumes a good measurement of customer profitability and different strategic and tactic retention measures to overcome the customer's movement.

In practice, a cellular operator can segment its customers by their profitability and focus retention management only on those profitable customer segments. The other way is, to score the entire customer data base with respect to propensity to churn and prioritize the retention effort based on profitability (life time value of customer) and churn propensity [6]. However, the telecom services industry in developing countries is yet to standardize a set of customer profitability measurements (e.g. current versus life-time value, business unit level versus corporate level value, number of accounts versus number of customers, loyalty versus profitability, etc.). This research assumes non-significant correlation between profitability and propensity to churn in order to simplify the modeling framework and to focuses on churn prediction based on available data mining techniques.

Burez and Van den Poel (2006) indicate that there are two types of targeted approaches to managing customer churn: reactive and proactive. When a company (cellular service provider) adopts a reactive approach, it waits until customers ask the company to cancel their service relationship. In this situation, the company may have to offer the customer an incentive to stay. On the other hand, when a company proceeds with a proactive approach, it tries to identify customers who are likely to churn to other service providers before they actually do so. The company then provides special promotion programs or incentives for these customers to avoid the customers from churning. Targeted proactive promotion programs have potential advantages of having lower incentive costs. However, these strategies may be very wasteful if churn predictions are inaccurate, because companies will be wasting money to incentivize customers who will not churn. Therefore, an accurate customer-churn prediction model is also critical for success of customer incentive programs [3].

### 2.2 Related Work
Building an effective customer churn prediction model using various techniques has become a decisive topic for business and academics in recent years. In order to understand how different studies have constructed their churn prediction models, this paper reviews some of current studies as shown in Table 1.

**Table 1. Related Literature about Customer Churn**

| Author | Data set | Prediction method |
|---|---|---|
| Coussement and Poel, 2008 [12] | Subscriber database | Support vector machines random forests logistic regression |
| Burez and Poel, 2006 [3] | Pay-TV company | Logistic regression and Markov chains random forests |
| Hung et al., 2006 [6] | Wireless telecom. company | Classification (decision tree, neural network) clustering (K-means) |
| Buckinx and Poel, 2005) [13] | Retailing dataset | Neural networks, logistic regression |
| Poel and Larivie're, 2004 [4] | European financial services company | Hazard model survival analysis |
| Kim and Yoon, 2004 [9] | Five mobile carriers in Korea | Logistic regression |
| Chiang et al., 2003 [14] | Network banking | Association rules |
| Wei and Chiu, 2002 [15] | Taiwan wireless telecommunications company | Classification (decision tree) |

Coussement and Van den Poel (2008) applied support vector machines (SVM) in a newspaper subscription churn context in order to construct a churn model. The customer churn prediction performance of the support vector machine model is benchmarked to logistic regression and random forecasts [12].

Burez and Van den Poel (2006) built a prediction model for European pay-TV company by using Markov chains and a random forest model benchmarked to a basic logistic model [3]. Other studies (e.g. Kim & Yoon, 2004) only used a binomial logistic regression to construct the prediction model [9].

According to research literature about customer churn, most of the related work focuses on using only one data mining method such as classification or clustering to mine the customer retention data. Few studies (Hung et al., 2006) applied more than one technology which was based on cluster analysis and classification [6]. From the review of the literature we can conclude that neural network can predict customer churn in different domains like Pay-TV [3], retail [13], banking [4] and finance [14].

This paper proposes a neural network (NN) based approach to predict customer churn in subscription of cellular wireless services.

## 3. THE PROPOSED NEURAL NETWORK (NN) BASED APPROACH

### 3.1 Artificial Neural Network

An ANN is a complex network that comprises a large set of simple nodes known as neural cells. ANN was proposed based on advanced biology research concerning human brain tissue and neural system, and can be used to simulate neural activities of information processing in the human brain [16]. ANN has the topological structures of information processing nodes that distribute information in parallel fashion. The mappings of inputs and estimated output responses are obtained via combinations of nonlinear transfer functions. We can make use of self-adaptive information pattern recognition methodology to analyze the training algorithms of the artificial neural networks using past experience, neural cells, memory and association, to process fuzzy, nonlinear, and noise-containing data without developing any mathematical models. The various algorithms of the neural networks training are Hebb, Delta, Kohonen, and BP computation. The mostly used BP computation algorithm is the error back propagation algorithm proposed by the PDP group of Rumelhart in 1985 [17].

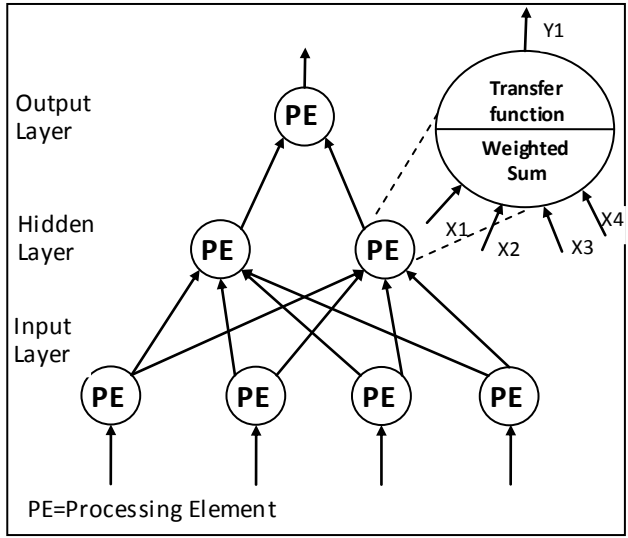

**Fig 1: Neural Network with One Hidden Layer**

Neural networks can be differentiated into single-layer perception and multilayer perception (MLP) network. The multilayer perception consists of multiple layers of simple, two state, sigmoid transfer function having processing element or neurons that interact by using weighted connections. In addition, the neural network contains one or more several intermediary hidden layers of neurons between the input and output layers. Such intermediary layers are called hidden layers and nodes embedded in these layers are called hidden nodes since these are not taking inputs directly from outside. A typical feed-forward multi-layer perceptron neural network consists of the input layer, output layer, and hidden layer (topological structure is as shown in Fig. 1). The neural networks adopting the error back propagation training algorithm are called BP networks, whose learning process comprises both backward and forward propagation. In forward processing, the sample signals will gradually progress through each layer with the sigmoid function $f(x) = 1/(1 + e^{-x})$.

The neural network cell i.e. neuron of each layer (input or hidden) only affects the status of the next neural cell. If the expected output signals cannot be obtained in the output layer, the weight values of each layer of the neural cells must be modified to keep the error minimum. Erroneous output signals will be backward from the source. Finally, the signal error will arrive in certain areas with repeated propagation thus the weights of the different layer neurons are modified.

The network is set at *n* layers and $y_j^n$, such that $y_j^n$ indicates the output with *n* layers and *j* nodes. If $y_j^0$ is equal to $x_j$, and *j* indicates the inputs. Let $W_{ij}^n$ be the connection weight between $y_i^{n-1}$ and $y_j^n$, then we get the threshold of $\theta_j^n$ of *n* layers and *j* nodes. The neural network learning algorithm includes the following steps:

1. Initialize node connection weights with random values and set other parameters.
2. Read in the input signal vector and the desired output. The signals progress at the networks with the following formula:
$$y_j^n = F(s_j^n) = F(\sum W_{ij}^n y_i^{n-1} + \theta_j^n) \quad (1)$$
It starts to calculate the *j* nodes of each layer with the output $y_j^n$ from the first layer through to the completion of calculation processing. F(*s*) represents one of the sigmoid transfer functions.
3. Compute the actual output via the calculations, working forward through the layers.
4. Compute the error. The error value of each node for the output layer is obtained from the different values between the real output and the required output ($D_j^k$) and is given by:
$$\delta_j^n = y_j^n(1 - y_j^n)(D_j^k - y_j^n) \quad (2)$$
The error value of each node for the previous (hidden) layers depends on the backward error propagation of each layer *(n = n, n-1... 1)* and is given by:
$$\delta_j^{n-1} = F(s_j^{n-1}) \sum W_{ij}^n \delta_i^n \quad (3)$$
5. Change the node connection weights by working backward from the output layer through the hidden layers which is done by the following formula:
$$W_{ij}^n(p+1) = W_{ij}^n(p) + \eta \delta_j^n y_i^{n-1} + \alpha[W_{ij}^n(p) - W_{ij}^n(p-1)] \quad (4)$$
$$\theta_j^n(p+1) = \theta_j^n(p) + \eta \delta_j^n + \alpha[\theta_j^n(p) - \theta_j^n(p-1)] \quad (5)$$
Where *p* indicates the iterative times (epoch) of the layers. The constant *η* indicates the learning rate and *α* indicates the momentum constant and their values can be from 0 to 1.

### 3.2 Dataset

This paper has used the churn data set from the UCI Repository of Machine Learning Databases at the University of California, Irvine [18]. The churn dataset deals with cellular service provider's customers and the data pertinent to the voice calls they make. Customers have a choice of service providers, or companies providing them with cellular network services. When these customers change cellular service provider they are said to churn which results in a loss of revenue for the previous cellular service provider. The telecommunications company concerned here has used all its databases like billing database, customer service database etc. and generated a list of pertinent records.

The neural network is implemented on Clementine data mining software package from SPSS, Inc [19]. The data set contains 20 variables worth of information about 2,427 customers, along with an indication of whether or not that customer churned (left the company). The variables are as follows-

- State: categorical variable, for the 50 states and the district of Columbia
- Account length: integer-valued variable for how long account has been active
- Area code: categorical variable
- Phone number: essentially a surrogate key for customer identification
- International Plan: dichotomous categorical having yes or no value
- Voice Mail Plan: dichotomous categorical variable having yes or no value
- Number of voice mail messages: integer-valued variable
- Total day minutes: continuous variable for number of minutes customer has used the service during the day
- Total day calls: integer-valued variable
- Total day charge: continuous variable based on foregoing two variables
- Total evening minutes: continuous variable for minutes customer has used the service during the evening
- Total evening calls: integer-valued variable
- Total evening charge: continuous variable based on previous two variables
- Total night minutes: continuous variable for storing minutes the customer has used the service during the night
- Total night calls: integer-valued variable
- Total night charge: continuous variable based on foregoing two variables
- Total international minutes: continuous variable for minutes customer has used service to make international calls
- Total international calls: integer-valued variable
- Total international charge: continuous variable based on foregoing two variables
- Number of calls to customer service: integer-valued variable

## 3.3 Training and Testing of Neural Network

When using neural networks to perform predictive modeling, the input layer contains all of the input fields or variables used to predict the outcome variable. The output layer contains an output field which is the target of the prediction. The input and output fields can be numeric or symbolic. In Clementine, symbolic fields are transformed into a numeric form (dummy or binary set encoding) before processing by the network. The hidden layer contains a number of neurons at which outputs from the previous layer combine. A network can have any number of hidden layers, although these are usually kept to a minimum to simplify the predictive model. All neurons in one layer of the network are connected to all neurons within the next layer while the neural network is learning the relationships between the data and results, it is said to be training. The Figure 2 provides details of implementation of neural network using Clementine 12.0.

Clementine provides two different classes of supervised neural networks, the Multi-Layer Perceptron (MLP) and the Radial Basis Function Network (RBFN). There are five different algorithms available within the Neural Net node of Clementine but this paper has used the widely applied Quick method. The Quick method use a feed-forward back-propagation network whose topology (number and configuration of nodes in the hidden layer) is based on the number and types of the input and output fields.

To prevent over-training problems that can occur within neural networks, randomly selected proportion of the training data is used to train the network. As the data pass repeatedly through the network, it is possible for the network to learn patterns that exist in the sample only and thus over-train. So the network may become too specific to the training sample data and loose its ability to generalize. The randomly selected proportion of the training data is used to train the network and once this proportion of data has made a complete pass through the network, the rest is reserved as a test set to evaluate the performance of the current neural network architecture.

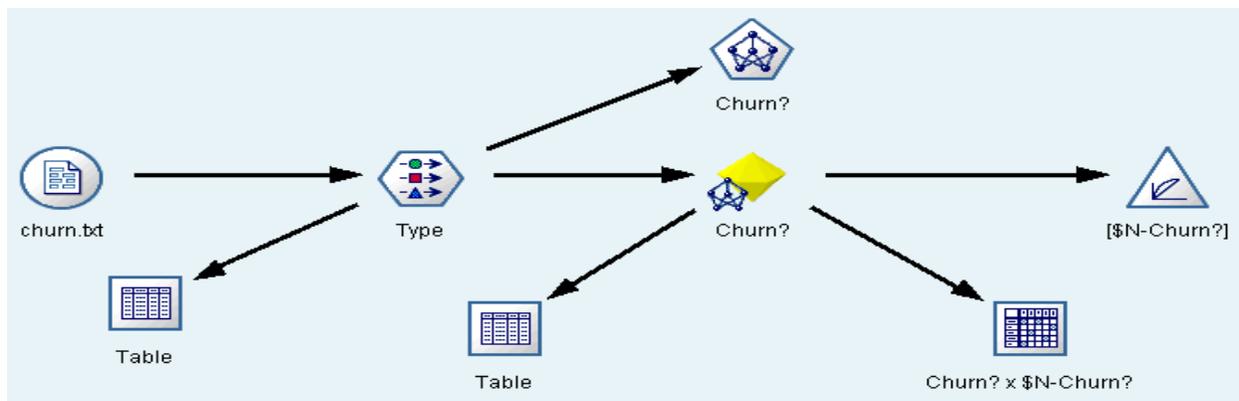

**Fig 2: The Implementation of Neural Network using Clementine 12.0**

## 4. RESULT AND DISCUSSION

The analysis section of the generated model displays information about the neural network. Figure 3 depicts the final model summary. The predicted accuracy for this neural network is 92.35%, indicating the proportion of the test set correctly predicted. The input layer is made up of one neuron per numeric or flag type field. The input variable named state and phone number were removed from the model since these variables were used for identification only. This study has experimented with multiple hidden layers in the neural network, containing three to seven neurons but the best results were obtained having one hidden layer with three neurons. The output layer contains two neurons corresponding to the two values of the output field (churn being true or false).

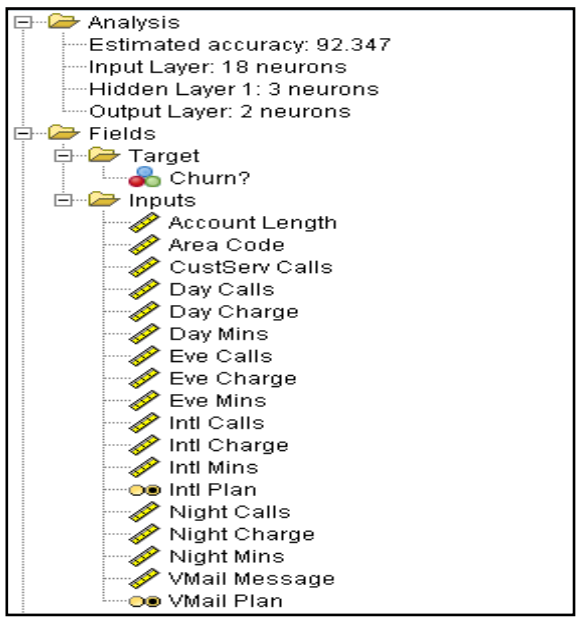

**Fig 3: The Generated Model**

Figure 4 represents relative importance of the input variables for doing sensitivity analysis of the generated model. The input fields are listed in descending order of relative importance. Importance values can range from 0.0 and 1.0, where 0.0 indicates unimportant and 1.0 indicates extremely important. In practice this figure rarely goes above 0.35. Here we see that Customer Service Calls is the most important field within this current network, and then followed by International Plan, Day Minutes and Day Charge. So we can conclude those customers who are frequently calling to customer service numbers, having international calling plan and spending much time in calling others during day time, may become potential churners.

The generated Neural Net calculates two new fields, $N-Churn and $NC- Churn, for every record in the input data base. The first represents the predicted Churn (true or false) and the second a confidence value for the prediction. The latter is only appropriate for symbolic outputs and will be in the range of 0.0 to 1.0, with the more confident predictions having values closer to 1.0. When predicting a symbolic field, it is valuable to produce a data matrix of the predicted values ($N-Churn) and the actual values (Churn) in order to study how they compare and where the differences are. Figure 5 represents matrix of actual (rows) and predicted (columns) churn of the proposed approach.

The model is predicting over 97% of the false churner means loyal customers, correctly but only 66% of true churners. If we wanted to correctly predict the loyal customers who will not churn at the expense of the other types this would appear to be a reasonable model. On the other hand, if we wanted to predict those customers that were going to cause the company a loss by churning, this model would only predict correctly in about two third of the instances.

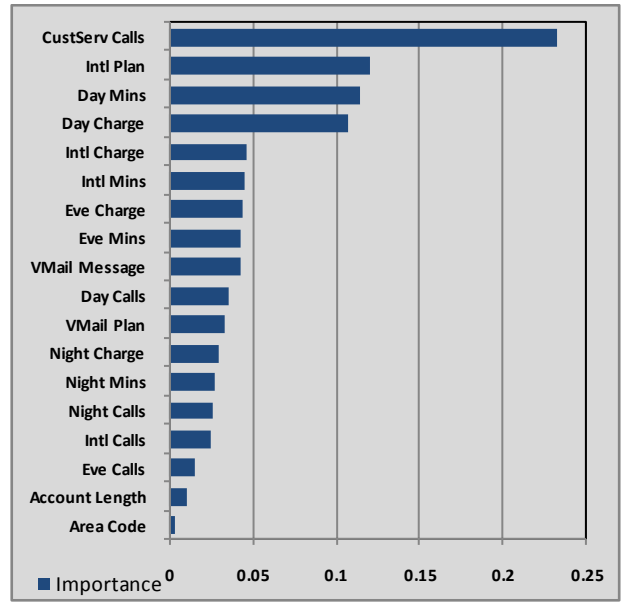

**Fig 4: The Relative Importance of the Input Variables**

| Churn? | | $N-Churn? | |
|---|---|---|---|
| | | False | True |
| | Count | 0 | 1 |
| | Row % | 0.000 | 100.000 |
| False | Count | 2038 | 50 |
| | Row % | 97.605 | 2.395 |
| True | Count | 114 | 224 |
| | Row % | 33.728 | 66.272 |

**Fig 5: Matrix of Actual (Rows) and Predicted (Columns) Churn**

## 5. CONCLUSION AND FUTURE RESEARCH

Churn prediction and management is crucial in liberalized cellular mobile telecom markets in developing countries. In order to be competitive in this market, cellular service providers have to be able to predict possible churners and take proactive actions to retain valuable loyal customers. Therefore, to build an effective and accurate customer churn prediction model, has become an important research problem for both academics and practitioners in recent years. This paper suggests that data mining techniques can be a promising solution for the customer churn management and we can establish an early-warning model for this non-steady-state customer system. The final model summary in this paper concludes that the model gives more than 92% overall accuracy for the prediction of the customer churn.

For future work, several issues can be considered. First, as the data pre-processing stage in data mining is a very important step for the final prediction model performance, the dimensionality reduction or feature selection step can be involved in addition to data reduction. Second, along with neural networks, other popular prediction techniques can be applied in combination, such as support vector machines, genetic algorithms, etc to develop hybrid models. Finally, the current methodology of churn prediction can be tested for other sectors like banking, insurance or air line and comparisons can be done for prediction accuracy.